\relax
\documentclass{article} 
\usepackage{ijcai19}  
\usepackage{times}  
\usepackage{helvet}  
\usepackage{courier}  
\usepackage{url}  
\usepackage{graphicx}  
\usepackage{linguex}
\usepackage{booktabs}
\usepackage{makeidx}  

\usepackage{enumitem}
\usepackage{enumerate}
\usepackage{xspace}
\usepackage{amsfonts}
\usepackage{amssymb}
\usepackage{amsmath,bm}
\usepackage{multirow}

\usepackage{caption}
\usepackage [autostyle, english = american]{csquotes}
\usepackage{algorithm}
\usepackage{algorithmic}
\MakeOuterQuote{"}

\makeatletter
\newcommand\thefontsize[1]{{#1 The current font size is: \f@size pt\par}}
\makeatother

\usepackage[colorlinks=true, allcolors=blue]{hyperref}

\newcommand{\mname}{\texttt{RDPD}\xspace}

\hypersetup{draft}
\begin{document}

\newcounter{sol} 
\setcounter{sol}{1} 

\title{\mname: Rich Data Helps Poor Data via Imitation}

\author{
Shenda Hong$^{1,2,5}$\and
Cao Xiao$^3$\and
Trong Nghia Hoang$^4$\and
Tengfei Ma$^4$\and \\
Hongyan Li$^{1,2}$\And 
Jimeng Sun$^5$\\
\affiliations
$^1$School of Electronics Engineering and Computer Science, Peking University, China\\
$^2$Key Laboratory of Machine Perception (Ministry of Education), Peking University, China\\
$^3$Analytics Center of Excellence, IQVIA, USA\\
$^4$IBM Research, USA\\
$^5$Department of Computational Science and Engineering, Georgia Institute of Technology, USA\\
\emails
hongshenda@pku.edu.cn,
cao.xiao@iqvia.com,
nghiaht@ibm.com,
Tengfei.Ma1@ibm.com,
lihy@cis.pku.edu.cn,
jsun@cc.gatech.edu
}

\maketitle
\begin{abstract}
In many situations, we need to build and deploy separate models in related environments with different data qualities. For example, an environment with strong observation equipments (e.g., intensive care units) often provides high-quality multi-modal data, which are acquired from multiple sensory devices and have rich-feature representations. On the other hand, an environment with poor observation equipment (e.g., at home) only provides low-quality, uni-modal data with poor-feature representations. To deploy a competitive model in a {\it poor-data} environment without requiring direct access to multi-modal data acquired from a {\it rich-data} environment, this paper develops and presents a knowledge distillation (KD) method (\mname) to enhance a predictive model trained on poor data using knowledge distilled from a high-complexity model trained on {\it rich, private} data. We evaluated \mname on three real-world datasets and shown that its distilled model consistently outperformed all baselines across all datasets, especially achieving the greatest performance improvement over a model trained only on low-quality data by $24.56\%$ on PR-AUC and $12.21\%$ on ROC-AUC, and over that of a state-of-the-art KD model by $5.91\%$ on PR-AUC and $4.44\%$ on ROC-AUC.

\end{abstract}

\section{Introduction}

Many {\it rich-data} environments encompass multiple data modalities. For example, multiple motion sensors in a lab can collect activity signals from various locations of a human body where signals generated from each location can be viewed as one modality. Multiple leads for Electrocardiogram (ECG) signals in hospital are used for diagnosing heart diseases, of which each lead is considered a modality. 
Multiple physiological signals are measured in Intensive Care Units (ICU) where each type of measure is a modality. A series of recent studies have confirmed that finding patterns among rich multimodal data can increase the accuracy of diagnosis, prediction, and overall performance of the deep learning models~\cite{doi:10.1093/jamia/ocy068,hong2017encase}.
 
Despite the promises that rich multimodal data bring us, in practice we have more {\it poor-data} environments with data from fewer modalities of limited quality. For example, unlike in a {\it rich-data} environment such as hospitals where patients place multiple electrons to collect 12-lead ECG signals, in everyday home monitoring devices often only measure lead I ECG signal from arms. Although deep learning models often perform well in {\it rich-data} environment, their performance on {\it poor-data} environment is less impressive due to limited data modality and lower quality~\cite{DBLP:journals/corr/abs-1708-04347}.

We argue that given both rich- and poor-data from similar contexts, the models built on rich multi-modal data can help improve the other model built on poor data with fewer modalities or even a single modality. 
For example, a heart disease detection model trained on 12 ECG channels in a hospital can help improve a similar heart disease detection model trained on ECG signals from a single-channel at home.

The recent development of mimic learning or knowledge distillation~\cite{hinton2015distilling,ba2014deep,lopez2015unifying} has provided a way of transferring information from a complex model (teacher model) to a simpler model (student model). Knowledge distillation or mimic learning essentially compresses the knowledge learned from a complex model into a simpler model that is much easier to deploy. However they often require the same data for teacher and student models. Domain adaptation techniques address the problem of learning models on some source data distribution that generalize to a different target distribution. Deep learning based domain adaptation methods have focused mainly on learning domain-invariant representations~\cite{Glorot:2011:DAL:3104482.3104547,Chen:2012:MDA:3042573.3042781,Bousmalis:2016:DSN:3157096.3157135}. However they often need to be trained jointly on source and target domain data and are therefore unappealing to the settings when the target data source is unavailable during training. 

In this paper, we propose \mname (Rich Data to Poor Data) to build accurate and efficient models for poor data with the help of rich data. In particular, \mname transfers knowledge from a teacher model trained on rich data to a student model operating on poor data by directly leveraging multimodal data in the training process. Given a teacher model along with attention weights learned from multimodal data, \mname is trained end-to-end for the student model operating on poor data to imitate the behavior (attention imitation) and performance (target imitation) of the teacher model. 

In particular, \mname jointly optimize the combined loss of attention imitation and target imitation. The loss of target imitation can utilize both hard labels from the data and soft labels provided by the teacher model. Here are the main contributions of this work:
\begin{itemize}
\item We formally define the learning task from rich data to poor data, which has many real-world applications including healthcare.
\item We propose \mname algorithm based on mimic learning, which takes a joint optimization approach to transfer knowledge learned by a teacher model using rich data to help improving a student model trained only on poor data. The resulting model is also much lightweight than the original teacher model and can be more easily deployed.
\item We show that \mname consistently outperformed all baselines across multiple datasets and achieve the greatest performance improvement over the Direct model trained on common features between rich and poor data by $24.56\%$ on PR-AUC and $12.21\%$ on ROC-AUC, and over the standard distillation model in~\cite{hinton2015distilling} by $5.91\%$ on PR-AUC and $4.44\%$ on ROC-AUC.
\end{itemize}

\section{Method}

\begin{figure}[t]
\centering
\includegraphics[width=\linewidth]{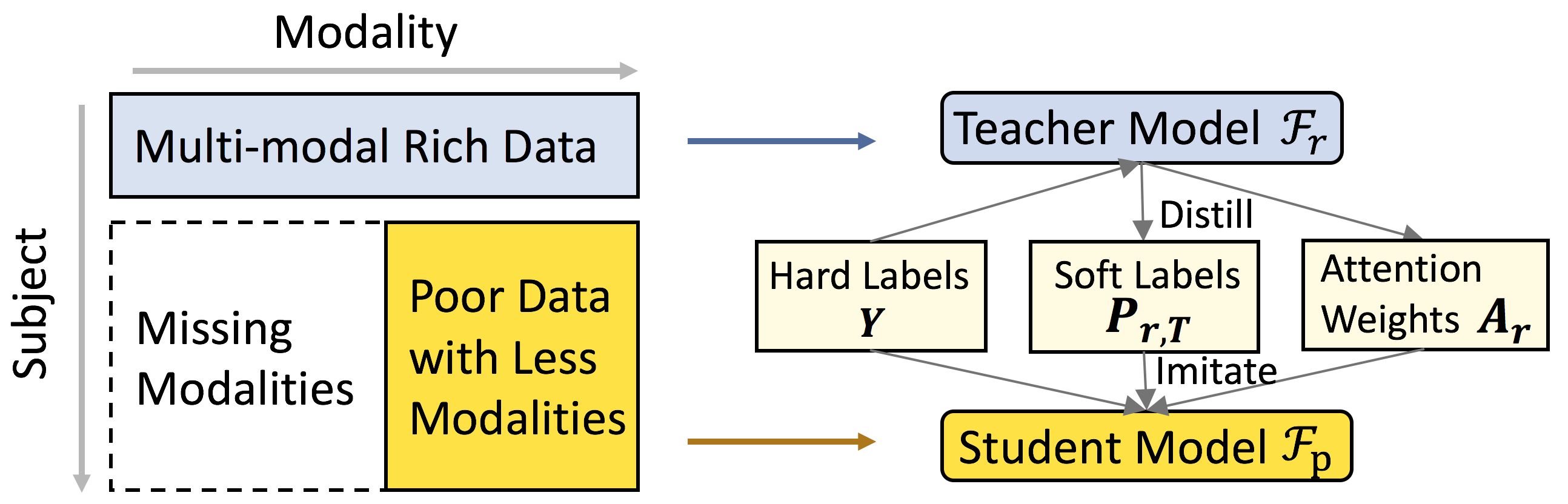}
\caption{The framework of \mname. Given teacher model along with attention weights learned from rich data, \mname trains the student model on poor data while imitating the behavior and performance of teacher model. In particular, \mname jointly optimize the combined loss of attention imitation (behavior) and target imitation (performance). The loss of target imitation also concerns both hard labels from data and soft labels provided by the teacher model.}
\label{fig:framework}
\end{figure}

In this section, we will first describe the task, and then introduce the design of \mname (shown in Figure~\ref{fig:framework}). 

\subsection{Task Description}

Consider data collected via continuous time series, given a teacher model trained from rich data environment, we want to teach a student model running on only poor data. And we hope the student model could benefit from the information contained in rich data via the teacher model, by imitating the teacher model in terms of \textbf{learning outcome} and the \textbf{learning process}. In this work, we call the former objective as \textbf{target imitation}, and the latter one as \textbf{behavior imitation}. Target imitation can be achieved by imitating the final predictions (i.e., soft labels) of the teacher model while behavior imitation can be achieved by imitating its attention weights over temporal time series. 

Mathematically, denote $\bm{X}_r$ as the multi-modal rich data with $D_r$ modalities that is available in training phase, and $\bm{X}_p$ as the poor data with $D_p$ modalities that is available in both training and testing phases. Here the modalities in $\bm{X}_p$ are a subset of $\bm{X}_r$, and $D_p < D_r$; $\bm{X}_p$ and $\bm{X}_r$ share the same labels $\bm{Y}$. Our task is to build a student model $\mathcal{F}_p$ which only takes $\bm{X}_p$ as input, and will benefit from knowledge transferred from $\bm{X}_r$.

\noindent\textbf{Overview.} For \mname, the student model trained on poor data  will imitate teacher model  trained on rich data and hard labels in both intermediate learning behavior and  final learning performance. The imitation of learning behavior is achieved by optimizing information loss from distribution of attention in student model  to distribution of attention in teacher model while the performance imitation is done by jointly optimizing hard label, soft label and and their trainable combination. In the following we will detail each step of \mname.

\subsection{Training Teacher Model}

Although \mname can be applied on time series in general, in this paper we only consider regularly sampled continuous time series $\bm{X_r}$ (e.g., sensor data). 
Assume a patient has time series from $D_r$ modalities, for time series in each modality with length $l$, we split $\bm{X_r} \in \mathbb{R}^{l \times D_r}$ into $M$ segments at length $S$, thus $l = M\times S$. We denote multi-modal segmented input time series as $ \bm{S_r} \in \mathbb{R}^{M \times S \times D_r}$. 

We applied stacked 1-D convolutional neural networks (CNN) on each segment and recurrent neural networks (RNN) across segments. Such a design has been demonstrated to be effective in many previous studies on multivariate time series modeling ~\cite{ordonez2016deep,choi2016convolutional}. In detail, we apply 1-D CNN with mean pooling on each segment $\bm{s_r}^{(j)} \in \mathbb{R}^{S \times D_r}, j=1,\cdots,M$ as given by Eq.~\ref{eq:1dcnn}. 
Parameters including number of filters, filter size and stride in CNN are shared among segments $\bm{s_r}^{(1)},\cdots,\bm{s_r}^{(M)}$, and vary across different datasets. Details are shown in the Experiment Setup section.
\begin{eqnarray}
\bm{h_r}^{(j)}  = Pooling(CNN_{1D}(\bm{s_r}^{(j)}))
\label{eq:1dcnn}
\end{eqnarray}
Then, we concatenate all convolved and pooled segments to get $\bm{H_r} = [\bm{h_r}^{(1)}, \cdots, \bm{h_r}^{(M)}]^T \in \mathbb{R}^{M \times K_r}$, where $K_r$ is the number of filters in $CNN_{1D}$.
Next we applied an RNN layer on $\bm{H_r}$ and denote the output as $\bm{Q_r}$ such that $\bm{Q_r} = RNN(\bm{H_r})$. And $\bm{Q_r} \in \mathbb{R}^{M \times U_r}$, where $U_r$ is the number of hidden units in RNN layer.  
Here we use the widely-applied self-attention mechanism ~\cite{lin2017structured} as it is a natural choice to get better results by taking advantage of the correlations or importance of segments. It also generates attention weights $\bm{A_r}$ that could represent teacher's behaviors on each segment. The attention weights are calculated by Eq.~\ref{eq:att_weight}. 
\begin{eqnarray}
\bm{A_r} = softmax(\bm{Q_r}\bm{W}) 
\label{eq:att_weight}
\end{eqnarray}
where $\bm{W} \in \mathbb{R}^{U_r \times 1}$, $\bm{A_r} \in \mathbb{R}^{M \times 1}$. We then multiplied the RNN output $\bm{Q_r}$ with corresponding attention weights $\bm{A_r}$. The weighted output $\bm{G_r}$ is given by Eq.~\ref{eq:att_rnn}. 
\begin{eqnarray}
\bm{G_r} &=& \bm{A_r}^{T}\bm{Q_r}
\label{eq:att_rnn}
\end{eqnarray}
where $\bm{G_r} \in \mathbb{R}^{1 \times U_r}$.
Finally, the weighted output $\bm{G_r}$ is further transformed by a dense layer with weights $\bm{W_d}\in \mathbb{R}^{U_r \times C}$ to output logits $\bm{O_r} \in \mathbb{R}^{1 \times C}$. 
\begin{eqnarray}
\bm{O_r} &=& \bm{G_r}\bm{W_d}
\label{eq:dense}
\end{eqnarray}

For simplicity, we can summarize from Eq.\ref{eq:1dcnn} to Eq.\ref{eq:dense} to represent the teacher model $\mathcal{F}_r$ as in Eq.~\ref{eq:ft}: $\mathcal{F}_r$ takes $\bm{X_r}$ as inputs and outputs logits $\bm{O_r}$ and attention weights $\bm{A_r}$.
\begin{eqnarray}
 \mathcal{F}_r(\bm{X_r})&=&\bm{A_r},\bm{O_r}\label{eq:ft}
\end{eqnarray}
The objective function of the teacher model measures prediction accuracy, and also provides knowledge to student model. 
Typically, $\bm{O_r}$ are transformed by softmax as final predicted probabilities, which can be used as distilled knowledge for student model to imitate. However, sharp distribution (e.g, hard labels)  will be less informative. To alleviate this issue, we follow the idea in ~\cite{hinton2015distilling} to produce more informative soft labels. Compared with hard label, the soft label imitation has much smoother probability distribution over classes, thus contains richer (larger entropy) informations.
Concretely, we modify classic softmax to $\mathcal{S}(x, T)$ by dividing original logits $\bm{O_r}$ with a predefined hyper-parameter $T$ (larger than 1). $T$ is usually referred to as Temperature. The modified softmax (shows $i$th soft probability) is given by Eq.~\ref{eq:modi_softmax} and the soft predictions are given by Eq.~\ref{eq:soft_target}.
\begin{eqnarray}
&\mathcal{S}(\bm{x}, T)_i& = \frac{\exp(x_i/T)}{\sum_j\exp(x_j/T)}\label{eq:modi_softmax}\\
&\bm{P_{r,T}}& = \mathcal{S}(\bm{O_r}, T) 
\label{eq:soft_target}
\end{eqnarray}
Finally, we use cross-entropy loss as prediction loss $\mathcal{L}_{teacher}$ (in Eq.~\ref{eq:opt_teacher}) to measure the difference between soft predictions $\bm{P_{r,T}} \in \mathbb{R}^{1 \times C}$ and ground truth $\bm{Y} \in \mathbb{R}^{1 \times C}$. We optimize teacher model via minimizing $\mathcal{L}_{teacher}$.
\begin{eqnarray}
\mathcal{L}_{teacher} = CrossEntropy(\bm{Y}, \bm{P_{r,T}})\label{eq:opt_teacher}
\end{eqnarray}

\subsection{Imitating Attentions and Targets}
\label{sec:imit}

After training teacher model on rich data, we now describe the imitation process for the student model. For attention imitation, we mean to mimic attention weights. For target imitation, the student model imitates the following components: 1) soft label that is more informative, 2) hard label that could improve performance (according to ~\cite{hinton2015distilling}), and 3) a trainable combination of both soft label and hard label. Again, we start with constructing the student model $\mathcal{F}_p$ using a CNN + RNN architecture, but with fewer filters in CNN and fewer hidden units in RNN. In our experiment, we roughly keep the proportion of hyper-parameters in teacher model to student model the same as the proportion of $D_r$ to $D_p$ using $K_r/K_p \approx D_r/D_p$, where $K_r$ and $K_p$ is the number of filters of CNN in teacher model and student model. Also, $U_r/U_p \approx D_r/D_p$, where $U_r$ and $U_p$ is the number of hidden units of RNN in teacher model and student model. Similar to Eq.\ref{eq:ft},  $\mathcal{F}_p$ takes $\bm{X_p}$ as inputs and outputs logits $\bm{O_p}$ and attention weights $\bm{A_p}$ as in Eq.~\ref{eq:fs}.
\begin{eqnarray}
 \mathcal{F}_p(\bm{X_p})&=&\bm{A_p}, \bm{O_p}
\label{eq:fs}
\end{eqnarray}

\subsubsection{Attention Imitation}
In Eq.\ref{eq:att_weight} we define attention weights to represent the influence of different time segments to the final predictions. We assume that the attention behavior of student model should resemble that of teacher model, and formulate the attention imitation as below. Given Eq.\ref{eq:ft} and Eq.\ref{eq:fs}, to enforce $\bm{A_p}$ and $\bm{A_r}$ to have similar distributions, we minimize the Kullback-Leibler (KL) divergence $\mathcal{L}_{att}$ given by Eq.~\ref{eq:loss1} to measure the information loss from distribution of attention in student model $\bm{A_p}$ to distribution of attention in teacher model $\bm{A_r}$.
\begin{eqnarray}
\mathcal{L}_{att} & =& D_{KL}(\bm{A_p}||\bm{A_r}) 
\label{eq:loss1}
\end{eqnarray}

\subsubsection{Imitating Hard Labels} For hard label imitation, we optimize the student model by minimizing cross entropy loss $\mathcal{L}_{hard}$ (in Eq.~\ref{eq:loss2}) that measures the difference between predicted target values and ground truth values $\bm{Y} \in \mathbb{R}^{1 \times C}$, where $C$ is the number of target classes, $\bm{P_{p,1}} = \mathcal{S}(\bm{O_p}, 1)$.
\begin{eqnarray}
\mathcal{L}_{hard} = CrossEntropy(\bm{Y}, \bm{P_{p,1}}),
\label{eq:loss2}
\end{eqnarray}

\subsubsection{Imitating Soft Labels} 
Given soft labels from $\mathcal{F}_r$, we produce soft predictions $\bm{P_{p,T}}$ by the same temperature $T$ on softmax in student model $\mathcal{F}_p$. Then, we optimize a cross entropy loss $\mathcal{L}_{soft}$ (in Eq.~\ref{eq:loss3}) that measures the differences between student and teacher. 
\begin{eqnarray}
\mathcal{L}_{soft} =  T^2 CrossEntropy(\bm{P_{r,T}}, \bm{P_{p,T}}) 
\label{eq:loss3}
\end{eqnarray}
Here, $\bm{P_{r,T}}$ is defined in Eq.\ref{eq:soft_target}. $\bm{P_{p,T}} = \mathcal{S}(\bm{O_p}, T)$. Since the magnitudes of gradients in Eq.\ref{eq:loss3} is scaled by $1/T^2$ as we divided logits by $T$, we should multiply the soft imitation loss by $T^2$ to keep comparable gradient during implementation. 

\subsubsection{Imitating Combined Label}
While hard labels provide certain prediction outcomes and soft labels provide probabilistic predictions, the two labels may even be opposite. To resolve the gap between the two labels, a reasonable solution is to combine them to yield uncertain prediction (probabilities of each class). Besides, while hard label imitation helps student model learn more information from data, soft label imitation transfer more knowledge from the teacher model (smoother distribution), each will lead to either more bias (comes from data) or more variance (comes from model). To leverage their benefits and make them complement each other, we propose to minimize a linear combination of hard labels and soft labels, denoted as $\bm{P_{p,comb}}$ as the follows:

\begin{eqnarray}
\bm{P_{p,comb}} =  \mathcal{S}(w_1 \bm{P_{p,1}} + w_2 \bm{P_{p,T}} + b, 1)
\label{eq:loss_comb}
\end{eqnarray}
where $w_1, w_2, b$ are learnable parameters. For the combined imitation, we also use cross entropy loss $\mathcal{L}_{comb}$ (in Eq.~\ref{eq:loss4}) to define the loss between $\bm{P_{p,comb}}$ and ground truth $\bm{Y}$.
\begin{eqnarray}
\mathcal{L}_{comb} = CrossEntropy(\bm{Y}, \bm{P_{p,comb}}) 
\label{eq:loss4}
\end{eqnarray}

\subsection{Joint Optimization}\label{sec:joint_opt}
Finally, for the student model to imitate attentions and targets simultaneously, we jointly optimize all loss functions above. 
Here, we simply summed them up to get the final objective function $\mathcal{L}_{student}$ given by Eq.~\ref{eq:loss}. We summarize the \mname method in Algorithm \ref{alg:alg}.
\begin{eqnarray}
\mathcal{L}_{student} =  \mathcal{L}_{att} + \mathcal{L}_{hard} + \mathcal{L}_{soft} +\mathcal{L}_{comb}
\label{eq:loss}
\end{eqnarray}

\begin{algorithm}[t]
\caption{\mname($\bm{X_r}, \bm{X_p}, \bm{Y}, T$)}
\label{alg:alg}
\begin{algorithmic}[1]
\STATE Build teacher model $\mathcal{F}_r$
\STATE Compute $\bm{A_r}, \bm{O_r} = \mathcal{F}_r(\bm{X_r})$
\STATE \quad \quad \quad \quad $\bm{P_{r,T}} = \mathcal{S}(\bm{O_r}, T)$
\STATE \quad \quad \quad \quad  $\mathcal{L}_{teacher} = CrossEntropy(\bm{Y}, \bm{P_{r,T}})$
\WHILE {not convergence}
\STATE Update weights of $\mathcal{F}_r$ by optimizing $\mathcal{L}_{teacher}$ using back-propagation
\ENDWHILE

\STATE Build student model $\mathcal{F}_p$
\STATE Compute $\bm{A_p}, \bm{O_p} = \mathcal{F}_p(\bm{X_p})$
\STATE \quad \quad \quad \quad $\bm{P_{p,T}} = \mathcal{S}(\bm{O_p}, T)$, $\bm{P_{p,1}} = \mathcal{S}(\bm{O_p}, 1)$
\STATE \quad \quad \quad \quad  $\bm{P_{p,comb}} =  \mathcal{S}(w_1 \bm{P_{p,1}} + w_2 \bm{P_{p,T}} + b, 1)$
\STATE \quad \quad \quad \quad  $\mathcal{L}_{att} = D_{KL}(\bm{A_p}||\bm{A_r})$
\STATE \quad \quad \quad \quad  $\mathcal{L}_{hard} = CrossEntropy(\bm{Y}, \bm{P_{p,1}})$
\STATE \quad \quad \quad \quad  $\mathcal{L}_{soft} = T^2 CrossEntropy(\bm{P_{r,T}}, \bm{P_{p,T}})$
\STATE \quad \quad \quad \quad  $\mathcal{L}_{comb} = CrossEntropy(\bm{Y}, \bm{P_{p,comb}})$
\STATE \quad \quad \quad \quad  $\mathcal{L}_{student} =  \mathcal{L}_{att} + \mathcal{L}_{hard} + \mathcal{L}_{soft} +\mathcal{L}_{comb}$
\WHILE {not convergence}
\STATE Update weights of $\mathcal{F}_p$ by optimizing $\mathcal{L}_{student}$ using back-propagation
\ENDWHILE
\end{algorithmic}
\end{algorithm}

\section{Experiments}

\subsection{Experiment Setup}

\subsubsection{Datasets} We used the following datasets in performance evaluation. Data statistics are summarized in Table \ref{table:data}. 

\paragraph{PAMAP2 Physical Activity Monitoring Data Set (PAMAP2)} ~\cite{reiss2012creating} contains 52 channels of sensor signals of 9 subjects wearing 3 inertial measurement units (IMU, 100Hz) and a heart rate monitor (HR, 9Hz). The average length of each subject is about 42k points. We down-sample the signals to 50 Hz and choose $S=64$ for experiment. We followed the "frame-by-frame analysis" in~\cite{reiss2012creating} to pre-process the time series with sliding windows of 5.12 seconds duration and 1 second stepping between adjacent windows.
The task is to classify signals into one of the $12$ different physical activities (e.g.,  walking, running, standing, etc.). In our experiment, we choose data of subject 105 for validation, subject 101 for testing, and others for training. 

\paragraph{The PTB Diagnostic ECG Database (PTBDB)} includes 15 channels of ECG signals collected from controls and patients of heart diseases ~\cite{bousseljot1995nutzung}. The database contains 549 records from 290 subjects. We down-sample the signals to 200 Hz and choose $S=500$ for experiment. 
Similar to PAMAP2, we pre-processed the data using "frame-by-frame analysis" with sliding windows of 10 seconds duration and 5 second stepping between adjacent windows. Our task is to classify signals into one of the 6 patient groups. In our experiment, we random divided the data into training (80\%), validation (10\%) and test (10\%) sets by subjects. 

\paragraph{The Medical Information Mart for Intensive Care (MIMIC-III)} is collected on over $58,000$ ICU patients at the Beth Israel Deaconess Medical Center (BIDMC) from June 2001 to October 2012 ~\cite{johnson2016mimic}. In our experiment, we focus on patients with following diseases: 1) acute myocardial infarction, 2) chronic ischemic heart disease, 3) heart failure, 4) intracerebral hemorrhage, 5) specified procedures complications, 6) lung diseases,7) endocardium diseases, and 8) septicaemia, in total $9,488$ subjects. In detail, we extract 6 vital sign time series of the first 48 hours including heart rate (HR), Respiratory Rate (RR), Blood Pressure mean, Blood Pressure systolic, Blood Pressure diastolic and SpO2. We resample the time series to 1 point per hour and choose $S=12$ for experiment. Our task is to classify vital sign series into one of the 8 diseases. In our experiment, we random divided the data into training (80\%), validation (10\%) and test (10\%) sets by patients.

\begin{table}[t]
\centering
\resizebox{1.0\linewidth}{!}{
\begin{tabular}{l|r|r|r}
\hline
                 & PAMAP2       & PTBDB       & MIMIC-III     \\
\hline
\# subjects      & 9            & 290         & 9,488     \\
\# classes       & 12           & 6           & 8         \\
\# attributes    & 52           & 15          & 6         \\
Total time series length        & 2,872,533    & 59,619,455  & 455,424   \\
\hline
\multirow{2}{*}{Sample Frequency} & 100 Hz (IMU)   & \multirow{2}{*}{1,000 Hz} & \multirow{2}{*}{1 per hour}   \\
& 9 Hz (HR) & &  \\
\hline
\end{tabular}
}
\caption{Statistics of Datasets}
\label{table:data}
\end{table}

\subsubsection{Evaluations and Implementation Details}

Performance was measured by the Area under the Receiver Operating Characteristic (ROC-AUC), Area under the Precision-Recall Curve (PR-AUC), and macro F1 score (macro-F1). ROC-AUC and PR-AUC are evaluated between predicted probabilities and ground truth. The PR-AUC is considered a better measure for imbalanced data with much more negative samples like our setting~\cite{davis2006relationship}. Macro-F1 is a commonly used with threshold $0.5$, which determine whether a given probability is predicted as $1$ (larger than threshold) or $0$ (smaller than threshold). 

Models are trained with the mini-batch of 128 samples for 200 iterations, which was a sufficient number of iterations for achieving the best performance for the classification task. The final model was selected using early stopping criteria on validation set. We then tested each model for 10 times using different random seeds, and report their mean values with standard deviation.
All models were implemented in PyTorch version 0.5.0., and trained with a system equipped with 64GB RAM, 12 Intel Core i7-6850K 3.60GHz CPUs and Nvidia GeForce GTX 1080.
All models were optimized using Adam ~\cite{adam}, with the learning rate set to 0.001. 
Our code is publicly available at \url{https://github.com/hsd1503/RDPD}. 

\subsubsection{Comparative Methods}
We will compare following methods: 
\begin{itemize}
\item \textbf{Teacher}: Teacher model is trained and tested on all channels. The model has better accuracy, a much heavier model architecture, and is only available for in-hospital setting where all channels of signals are available. It serves as an empirical upper bound of performance. 
\item \textbf{Direct}: Direct model is build on the partially observed data using RCNN, without attention imitation and soft label imitation. This model is equivalent to $\mathcal{L} =  \mathcal{L}_{hard}$. 
\item \textbf{Knowledge Distillation (KD)}: KD ~\cite{hinton2015distilling} model is constructed on the partially observed data, with soft label imitation and hard label imitation. This model is equivalent to $\mathcal{L} =  \mathcal{L}_{hard} + \mathcal{L}_{soft}$. 
\item Ours including \textbf{ $\mname_{\bm{r1}}$}: The reduced version of \mname without attention imitation. And the objective function would be  $\mathcal{L} =  \mathcal{L}_{comb} + \mathcal{L}_{hard} + \mathcal{L}_{soft}$. \textbf{$\mname_{\bm{r2}}$}: The reduced version of \mname without combined labels. This model is equivalent to KD model with attention imitation. And the objective function would be $\mathcal{L} =  \mathcal{L}_{att} + \mathcal{L}_{hard} + \mathcal{L}_{soft}$. \textbf{\mname}: Our whole model contains all proposed imitations. Using $\mathcal{L} =  \mathcal{L}_{att} + \mathcal{L}_{hard} + \mathcal{L}_{soft} +\mathcal{L}_{comb}$ as objective function.  
\end{itemize}

For all models, we use 1 layer 1-D CNN and 1 layer Bi-directional LSTM. In teacher model, for PAMAP2, the number of filters is set to 64, filter size is set to 8, stride is set to 4 and the number of hidden units is set to 32. For PTBDB, they are set to 128, 32, 8, 32 respectively. For MIMIC-III, they are set to 64, 4, 2, 32 respectively. In \mname and compared baselines, since they have less input modalities, they have smaller number of CNN filters and RNN hidden units which is set proportionally as introduced before. However, the data length remains the same, so their filter size and stride keep unchanged. $T$ is set to 5 for PAMAP2 and PTBDB, and set to 2.5 for MIMIC-III. 

\subsection{Results}

\subsubsection{Classification Performance}
We compared the results of \mname against other baselines and the reduced version of \mname in Table \ref{table:compare_pamap} (PAMAP2 dataset), Table \ref{table:compare_ptbdb} (PTBDB dataset) and Table \ref{table:compare_mimic} (MIMIC-III dataset). \mname outperformed other methods (except Teacher) in most cases and demonstrated the proposed attention imitation and target imitation successfully improved performance of student model. The teacher model performs best among all methods since it is trained using a full datasets with multiple modalities. It serves an empirical upper bound of the performance. In Table \ref{table:compare_ptbdb}, \mname works better than its reduced version in PR-AUC and F1-score but not ROC-AUC. The reason is that classes in PTBDB dataset is very imbalanced, some occasional samples in rare classes distort the final result. 

\begin{table}[t]
\centering
\resizebox{\linewidth}{!}{
\begin{tabular}{l|l|cccc}
\hline
Data                                                & Method  & ROC-AUC             & PR-AUC              & macro-F1             \\
\hline
All                     & Teacher & 0.928 $\pm$ 0.014 & 0.708 $\pm$ 0.039 & 0.608 $\pm$ 0.045  \\
\hline\hline
\multirow{4}{*}{Wrist}  & Direct  & 0.800 $\pm$ 0.032 & 0.452 $\pm$ 0.051 & 0.376 $\pm$ 0.049  \\
\multirow{4}{*}{}       & Distill & 0.825 $\pm$ 0.020 & 0.469 $\pm$ 0.052 & 0.380 $\pm$ 0.060  \\
                        & $\mname_{r1}$  & 0.837 $\pm$ 0.025 & 0.491 $\pm$ 0.037 & 0.406 $\pm$ 0.053  \\
                        & $\mname_{r2}$  & 0.836 $\pm$ 0.018 & 0.478 $\pm$ 0.038 & 0.401 $\pm$ 0.049  \\                                             & \mname  & \textbf{0.838 $\pm$ 0.012} & \textbf{0.491 $\pm$ 0.045} & \textbf{0.425 $\pm$ 0.057}  \\
\hline
\multirow{4}{*}{Chest}  & Direct  & 0.836 $\pm$ 0.035 & 0.519 $\pm$ 0.065 & 0.449 $\pm$ 0.069  \\
\multirow{4}{*}{}       & Distill & 0.868 $\pm$ 0.025 & 0.575 $\pm$ 0.043 & 0.486 $\pm$ 0.065  \\
                        & $\mname_{r1}$   & 0.872 $\pm$ 0.028 & 0.605 $\pm$ 0.030 & 0.518 $\pm$ 0.037  \\
                        & $\mname_{r2}$   & 0.879 $\pm$ 0.027 & 0.600 $\pm$ 0.051 & 0.478 $\pm$ 0.048  \\
                        & \mname   & \textbf{0.883 $\pm$ 0.016} & \textbf{0.609 $\pm$ 0.052} & \textbf{0.529 $\pm$ 0.051}  \\
\hline
\multirow{4}{*}{Ankle}  & Direct  & 0.811 $\pm$ 0.035 & 0.513 $\pm$ 0.065 & 0.405 $\pm$ 0.080  \\
\multirow{4}{*}{}       & Distill & 0.901 $\pm$ 0.015 & 0.621 $\pm$ 0.044 & 0.492 $\pm$ 0.070  \\
                        & $\mname_{r1}$      & 0.889 $\pm$ 0.021 & 0.581 $\pm$ 0.071 & 0.443 $\pm$ 0.095  \\
                        & $\mname_{r2}$      & 0.904 $\pm$ 0.019 & 0.629 $\pm$ 0.041 & 0.473 $\pm$ 0.069  \\
                        & \mname   & \textbf{0.910 $\pm$ 0.014} & \textbf{0.639 $\pm$ 0.030} & \textbf{0.511 $\pm$ 0.033}  \\
\hline
\end{tabular}
}
\caption{Performance comparison on PAMAP2 dataset. The task is multi-class classification (12 classes). All contains 52 channels, Wrist contains 17 channels signals of 1 IMU over the wrist on the dominant arm, Chest contains 17 channels signals of 1 IMU on the chest, Ankle contains 17 channels signals of 1 IMU on the dominant side's ankle. }
\label{table:compare_pamap}
\end{table}

\begin{table}[t]
\centering
\resizebox{\linewidth}{!}{
\begin{tabular}{l|l|cccc}
\hline
Data                       & Method  & ROC-AUC             & PR-AUC              & macro-F1             \\
\hline
All                        & Teacher & 0.737 $\pm$ 0.035 & 0.293 $\pm$ 0.018 & 0.288 $\pm$ 0.028 
  \\
\hline\hline
\multirow{4}{*}{Lead I}    & Direct  & 0.701 $\pm$ 0.023 & 0.279 $\pm$ 0.017 & 0.164 $\pm$ 0.020   \\
\multirow{4}{*}{}          & Distill & 0.676 $\pm$ 0.045 & 0.282 $\pm$ 0.022 & 0.217 $\pm$ 0.016   \\
                           & $\mname_{r1}$  & 0.677 $\pm$ 0.036 & 0.255 $\pm$ 0.029 & 0.139 $\pm$ 0.027   \\
                           & $\mname_{r2}$  & \textbf{0.707 $\pm$ 0.073} & 0.282 $\pm$ 0.044 & 0.218 $\pm$ 0.024   \\
                           & \mname  & 0.706 $\pm$ 0.075 & \textbf{0.293 $\pm$ 0.025} & \textbf{0.218 $\pm$ 0.019}   \\
\hline
\end{tabular}
}
\caption{Performance comparison on PTBDB dataset. The task is multi-class classification (6 classes). All contains 15 channels of ECG signals. Lead I contains single channel Lead I ECG signal, which is usually generated by mobile devices. }
\label{table:compare_ptbdb}
\end{table}

\begin{table}[t]
\centering
\resizebox{\linewidth}{!}{
\begin{tabular}{l|l|cccc}
\hline
Data                                                & Method  & ROC-AUC             & PR-AUC              & macro-F1             \\
\hline
All                  & Teacher & 0.696 $\pm$ 0.011 & 0.281 $\pm$ 0.009 & 0.256 $\pm$ 0.012 
  \\
\hline\hline
\multirow{4}{*}{BP}  & Direct  & 0.610 $\pm$ 0.016 & 0.204 $\pm$ 0.011 & 0.149 $\pm$ 0.013   \\
\multirow{4}{*}{}    & Distill & 0.611 $\pm$ 0.013 & 0.206 $\pm$ 0.007 & 0.150 $\pm$ 0.005   \\
                     & $\mname_{r1}$  & 0.607 $\pm$ 0.012 & 0.203 $\pm$ 0.003 & 0.148 $\pm$ 0.003   \\
                     & $\mname_{r2}$  & 0.613 $\pm$ 0.020 & 0.205 $\pm$ 0.009 & 0.147 $\pm$ 0.007   \\
                     & \mname    & \textbf{0.614 $\pm$ 0.018} & \textbf{0.207 $\pm$ 0.010} & \textbf{0.150 $\pm$ 0.006}   \\
\hline
\multirow{4}{*}{HR}  & Direct  & 0.556 $\pm$ 0.019 & 0.176 $\pm$ 0.013 & 0.089 $\pm$ 0.042   \\
\multirow{4}{*}{}    & Distill & 0.564 $\pm$ 0.021 & 0.175 $\pm$ 0.012 & 0.109 $\pm$ 0.030   \\
                     & $\mname_{r1}$  & 0.566 $\pm$ 0.010 & 0.178 $\pm$ 0.004 & 0.132 $\pm$ 0.005   \\
                     & $\mname_{r2}$  & 0.571 $\pm$ 0.011 & 0.176 $\pm$ 0.008 & 0.123 $\pm$ 0.016   \\
                     & \mname  & \textbf{0.581 $\pm$ 0.014} & \textbf{0.182 $\pm$ 0.004} & \textbf{0.130 $\pm$ 0.010}   \\
\hline
\multirow{4}{*}{RR}  & Direct  & 0.570 $\pm$ 0.019 & 0.176 $\pm$ 0.012 & 0.109 $\pm$ 0.039   \\
\multirow{4}{*}{}    & Distill & 0.614 $\pm$ 0.023 & 0.201 $\pm$ 0.009 & 0.162 $\pm$ 0.015   \\
                     & $\mname_{r1}$  & 0.611 $\pm$ 0.014 & 0.202 $\pm$ 0.007 & 0.160 $\pm$ 0.016   \\
                     & $\mname_{r2}$  & 0.614 $\pm$ 0.017 & 0.205 $\pm$ 0.006 & 0.169 $\pm$ 0.010   \\
                     & \mname  & \textbf{0.619 $\pm$ 0.022} & \textbf{0.207 $\pm$ 0.008} & \textbf{0.169 $\pm$ 0.007}   \\
\hline
\end{tabular}
}
\caption{Performance comparison on MIMIC-III dataset. The task is multi-class classification (8 classes). All contains 6 channels of patient vital signs. BP contains blood pressure systolic and blood pressure diastolic, which is usually monitors by house sphygmomanometer. HR is heart rate, RR is respiration rate.  }
\label{table:compare_mimic}
\end{table}

\subsubsection{Reduction of Model Complexity}

We analyzed model complexity by comparing model size of the teacher model and \mname. Table \ref{table:complexity} shows that the model size of \mname is only $6-7\%$ of the model size of teacher model. According to experimental settings and previous results, other methods have comparable model size with our approach, but their performance are worse. In real world applications such as mobile health or ICU real-time modeling, it is very important that \mname can achieve both lighter in model and better in performance. 
\begin{table}[t]
\centering
\resizebox{0.75\linewidth}{!}{
\begin{tabular}{l|ccc}
\hline
Model   & PAMAP2 & PTBDB & MIMIC-III \\
\hline
Teacher & 118.3k & 335.0k & 60.2k    \\
\mname & 8.2k & 19.8k & 4.0k       \\
\hline
\end{tabular}
}
\caption{Model complexity comparison, the table shows number of parameters of each model. }
\label{table:complexity}
\end{table}
\subsubsection{Evaluation against Size of Rich Data}

\begin{figure}[t]
\centering
\begin{tabular}{cc}
\hspace{-3mm}\includegraphics[width=0.5\linewidth]{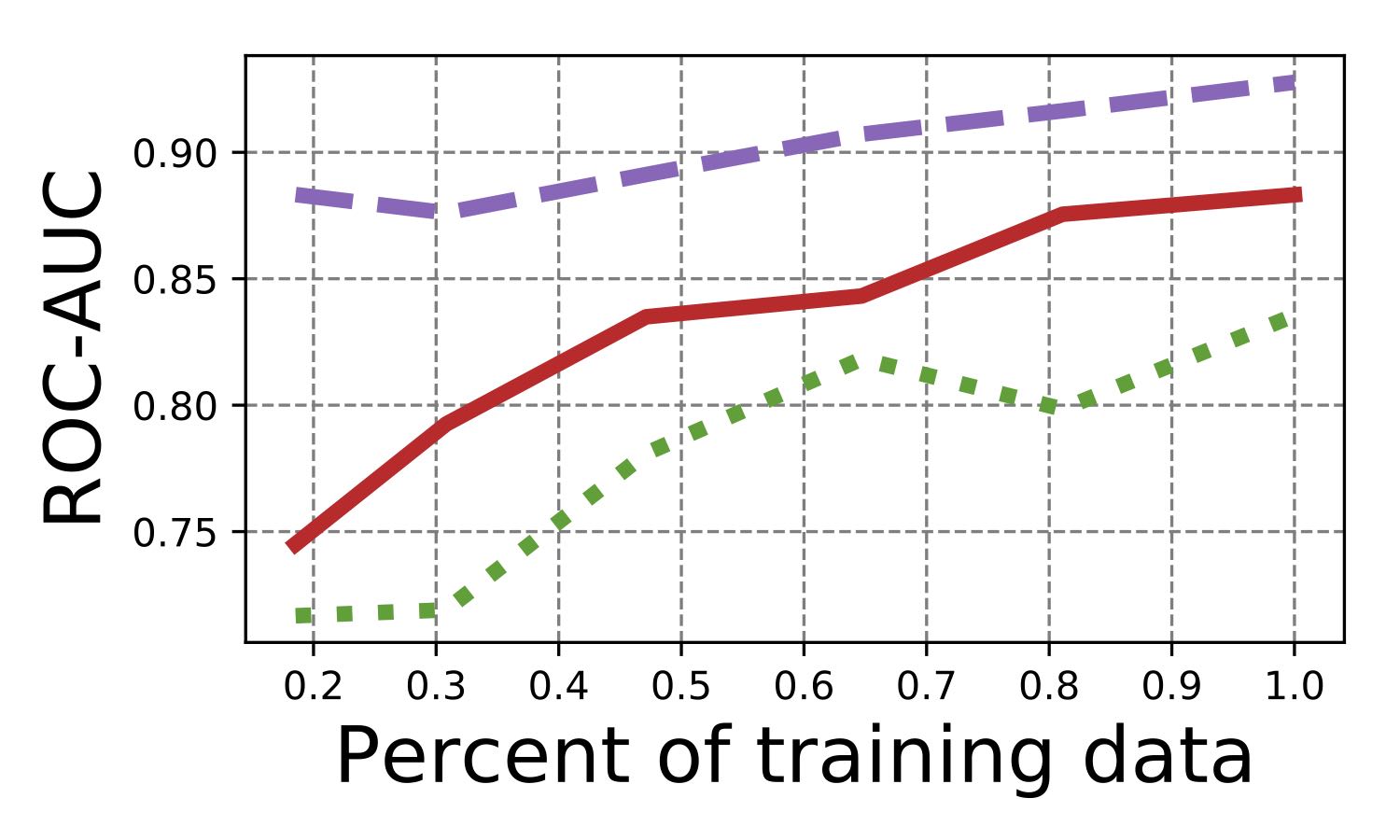}
&\hspace{-3mm}\includegraphics[width=0.5\linewidth]{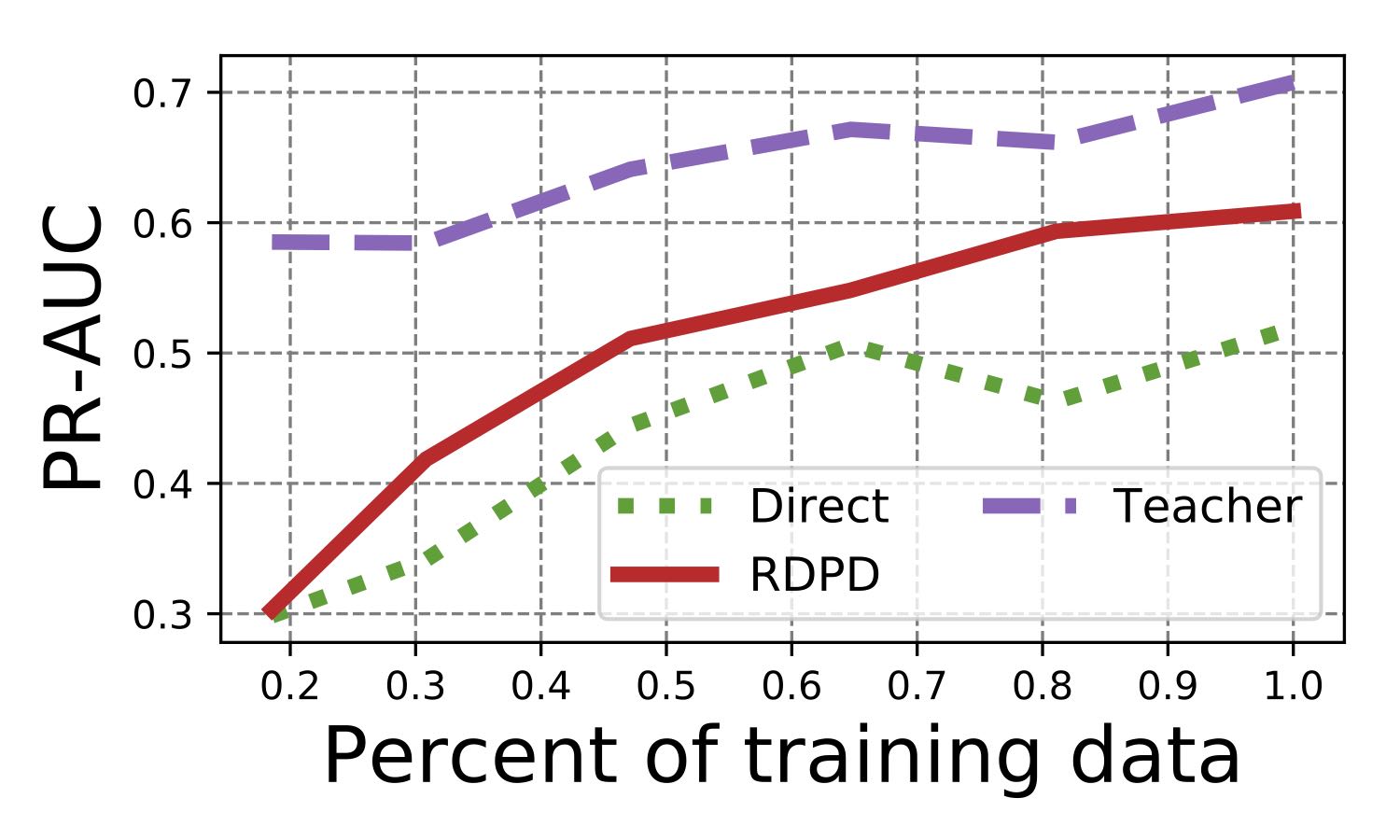}
\end{tabular}
\caption{Performance comparison of training data size using PAMAP2 dataset. }
\label{fig:size}
\end{figure}

We evaluated the dependency of size of rich data. We used the same validation and test data, but scaled down the size of rich data in training. Figure~\ref{fig:size} shows \mname outperformed baselines even we have few rich data, and would perform better as we got more rich data. This demonstrated the efficacy of \mname in extracting useful knowledge from rich data and teaching student even under {rich-data} insufficiency.

\subsubsection{Evaluation for the Setting of Low Quality Poor-data}
\begin{figure}[t]
\centering
\begin{tabular}{cc}
\hspace{-3mm}\includegraphics[width=0.5\linewidth]{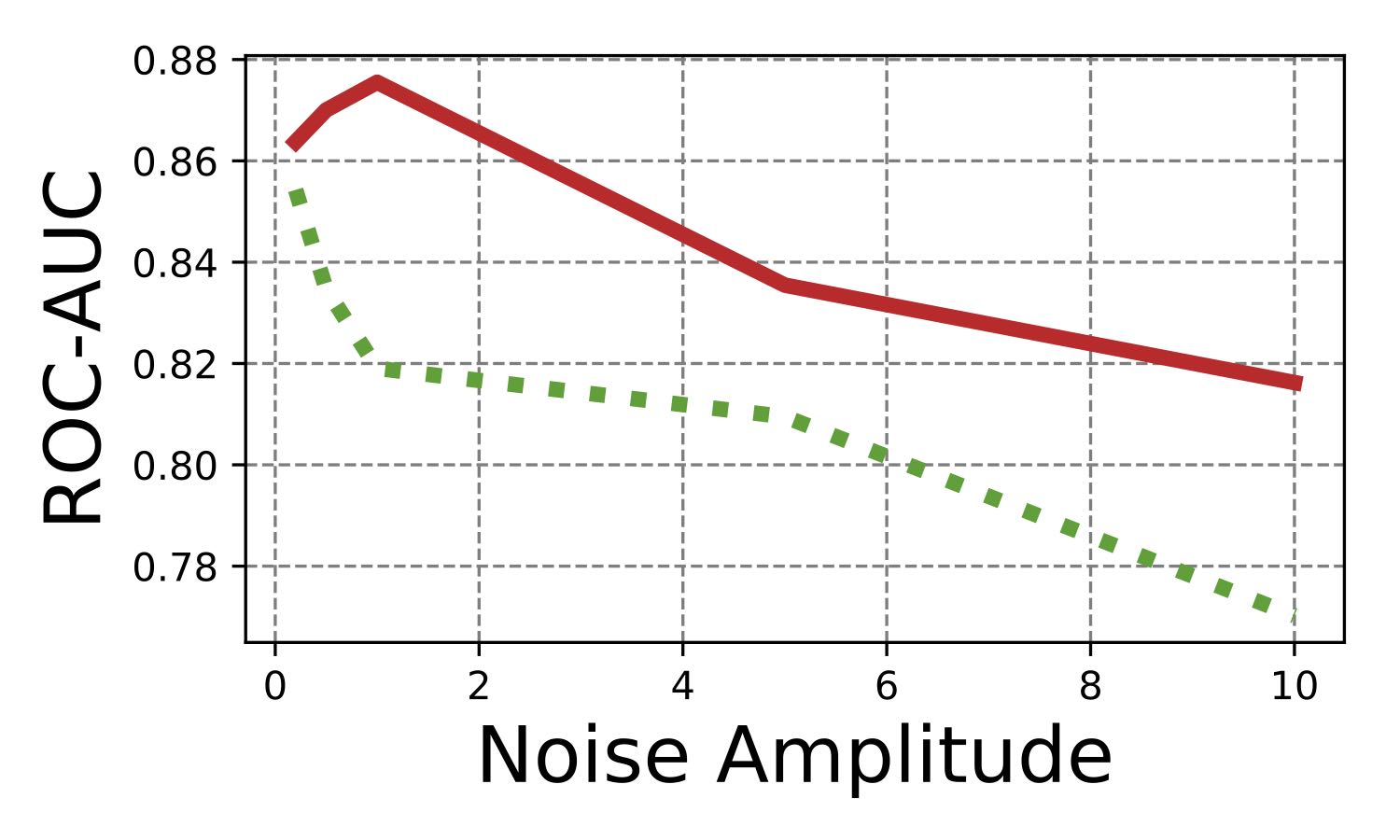}
&\hspace{-3mm}\includegraphics[width=0.5\linewidth]{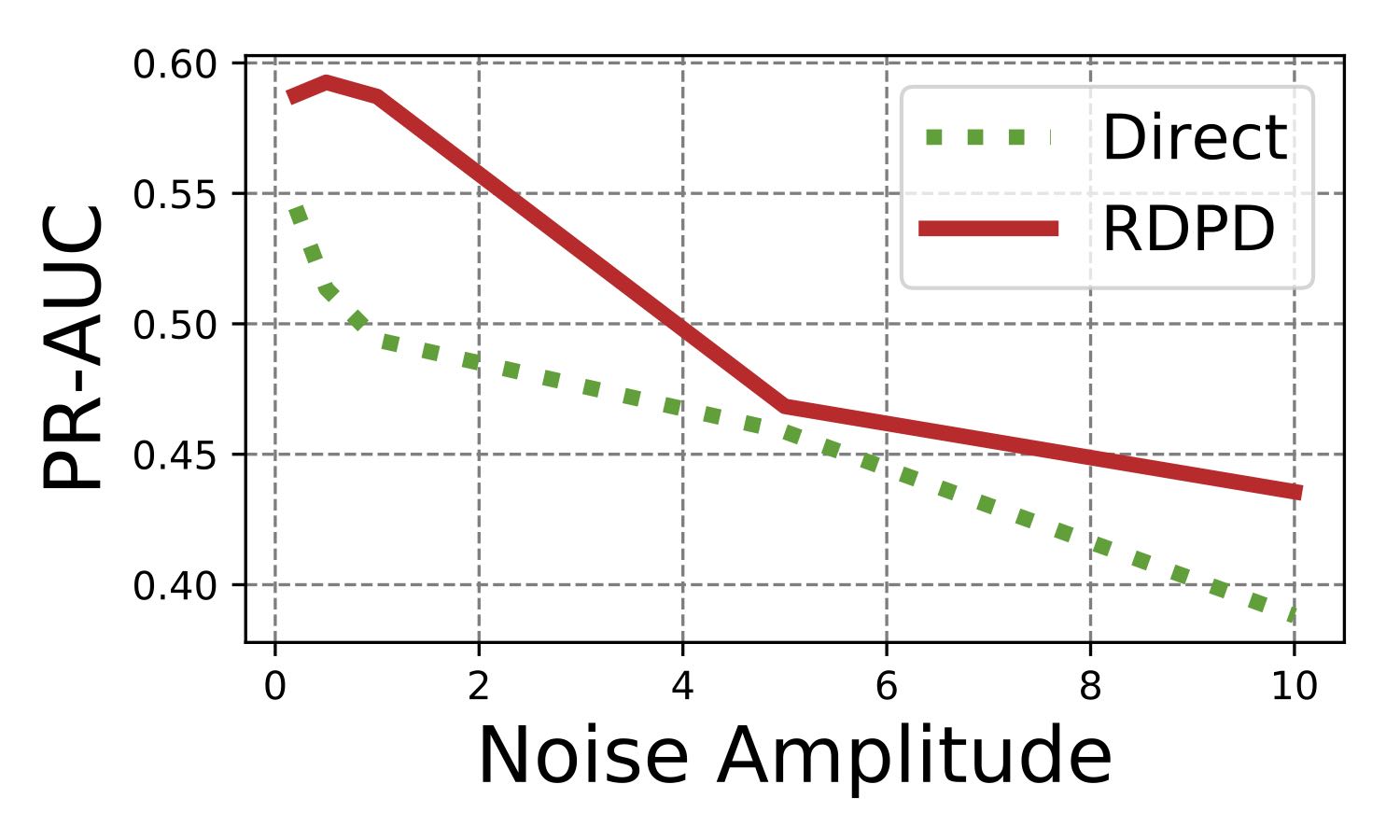}
\end{tabular}
\caption{Performance comparison of noise amplitude using PAMAP2 dataset. }
\label{fig:noise}
\end{figure}

Here we also assess how much benefit the multi-modality data can bring us from low quality poor-data. We performs experiments by adding different level of noise to the entire single modality. The approach of adding noise is: 
$\bm{x'} =\bm{x} \oplus amp*random\_normal(-1,1)$, 
where $\bm{x}$ is the original data and $\bm{x'}$ is the noise interfered data, $\oplus$ is element-wise add, $amp$ is the parameter to control the noise amplitude. From Figure \ref{fig:noise}, we can see with the increasing amplitude of noise, the performance of both Direct and \mname decrease. However, \mname still works better than Direct due to knowledge transfer from Teacher model. 

\section{Related Work}

\paragraph{Domain adaptation} Domain adaptation techniques address the problem of learning models on some source data distribution that generalize to a different target distribution. Deep learning based domain ad aptation methods have focused mainly on
learning domain-invariant representations. For example, 
~\cite{Glorot:2011:DAL:3104482.3104547} and ~\cite{Chen:2012:MDA:3042573.3042781} stacked Denoising
Auto-encoders (SDA) and marginalized SDA respectively  to extract meaningful representations. ~\cite{Ganin:2016:DTN:2946645.2946704} added a Gradient Reversal Layer that hinders the model’s ability to discriminate between domains. Moreover, ~\cite{Zhou2016BiTransferringDN}
transferred the source examples to the target domain and vice versa using BiTransferring Deep Neural Networks, while ~\cite{Bousmalis:2016:DSN:3157096.3157135} propose Domain Separation Networks. However they need to be trained jointly on source and target domain data and are therefore unappealing to the settings where both data are available.

\paragraph{Knowledge Distillation} Knowledge Distillation ~\cite{hinton2015distilling} or mimic learning ~\cite{ba2014deep} are a family of approaches that aim to transfer the predictive power from more accurate deep models ("teacher model") to smaller models ("student model") like shallow neural networks ~\cite{hinton2015distilling}, soft decision tree ~\cite{frosst2017distilling} via training smaller models on soft labels learned from deep models. It has been widely used in model compression ~\cite{sau2016deep}, omni-supervised learning ~\cite{radosavovic2017data}, fast optimization, network minimization and transfer learning ~\cite{yim2017gift}. Extensions of knowledge distillation unifies distillation and privileged information into generalized distillation framework to learn from multiple machines and data representations ~\cite{lopez2015unifying}. 
The performance of distilled shallow neural networks are often better than models that are directly built on training data. The biggest difference between our approach and knowledge distillation is that, 
knowledge distillation focus on transfer powerful predictions ability of teacher to student model, 
while our approach is designed to transfer both behaviors and predictions from rich data modalities to poor data (a single modality). 

\paragraph{Attention Transfer} Attention mechanism ~\cite{bahdanau2014neural} was proposed to improve performance of machine translation by paying more attention on relevant parts of the data. Recently, there are several works studying attention transfer ~\cite{Zagoruyko2017AT,huang2017like} to enhance shallow neural networks. The goal was achieved by learning similar attention models in smaller neural networks, then defining attention as gradient with respect to the input ~\cite{Zagoruyko2017AT} or use regularization term ~\cite{huang2017like} to make two models have similar attention weights. Attention transfer has been used in video recognition from web images ~\cite{li2017attention}, cross-domain sentiment classification ~\cite{li2018hierarchical} and so on. The biggest difference between our approach and attention transfer is that attention transfer is used for model compression on one dataset, while our approach is used to transfer across datasets of very different data modalities.

\section{Conclusion}

In this paper we proposed to leverage the power of rich data to improve the learning from poor data with \mname. \mname learns end-to-end for the student model built on poor data to imitate the behavior (attention imitation) and performance (target imitation) of teacher model by jointly optimizing the combined loss of attention imitation and target imitation. We evaluated \mname across multiple datasets and demonstrated its promising utility and efficacy. Future extension of \mname includes considering modeling static meta information as one modality, and learning from less labels.

\section*{Acknowledgements}
This work was supported by the National Science Foundation, award IIS-1418511, CCF-1533768 and IIS-1838042, the National Institute of Health award 1R01MD011682-01 and R56HL138415.

\bibliographystyle{named}
\bibliography{ref}

\ifthenelse{\value{sol}=1}{

\clearpage

\appendix{\mname: Rich Data Helps Poor Data via Imitation}

\section{Case Study}

To further analyze the effects and reasons that how \mname improves performance. We choose several subjects and compared their attention weights and prediction results learned by different methods.

\begin{figure}[ht]
\centering
\includegraphics[width=\linewidth]{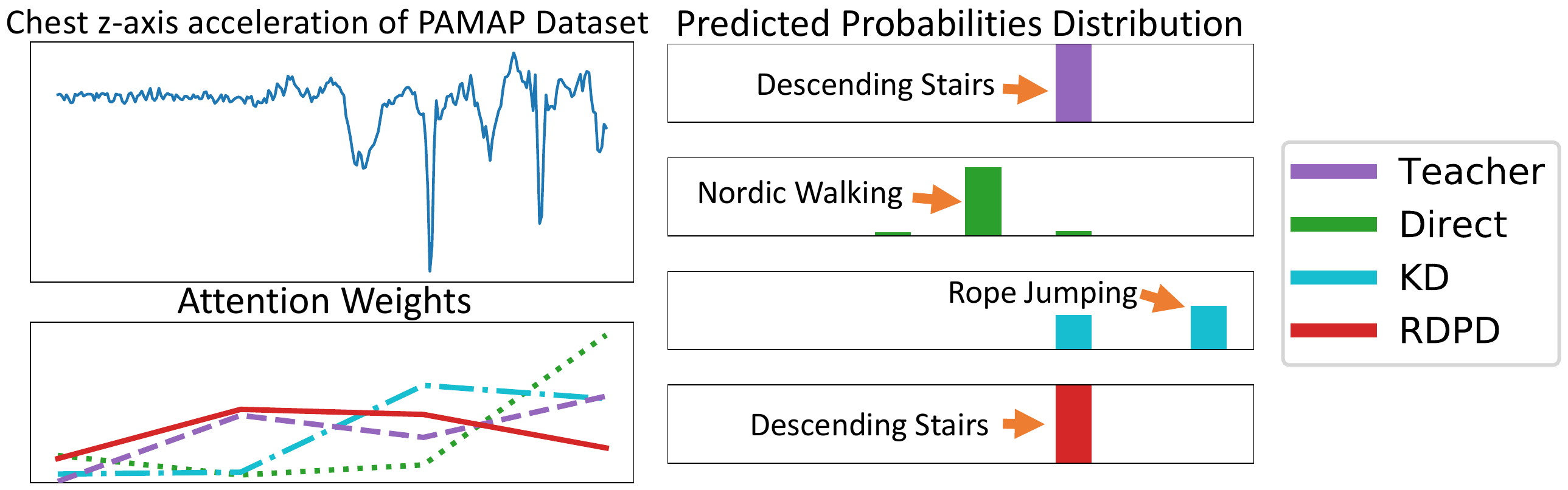}
\caption{An example on PAMAP2 data.}
\label{fig:case_pamap}
\end{figure}

Fig.~\ref{fig:case_pamap} shows an example from PAMAP2. Both \mname and Teacher correctly predict the subject is descending stairs, while Direct predicts the subject is doing Nordic walking and KD gives prediction of rope jumping. Fig.~\ref{fig:case_pamap} left upper plots the  z-axis acceleration from chest sensors, which shows the vertical acceleration of the whole body. Although these activities are similar, the change between walking on the floor and stairs distinguish the data from descending stairs with Nordic walking and rope jumping. Teacher model provide correct prediction  by looking more channels, thus gives more attention in the vertical acceleration. \mname also predicts correctly since it imitates the attentions from Teacher as shown in Figure \ref{fig:case_pamap} left bottom.

\begin{figure}[ht]
\centering
\includegraphics[width=\linewidth]{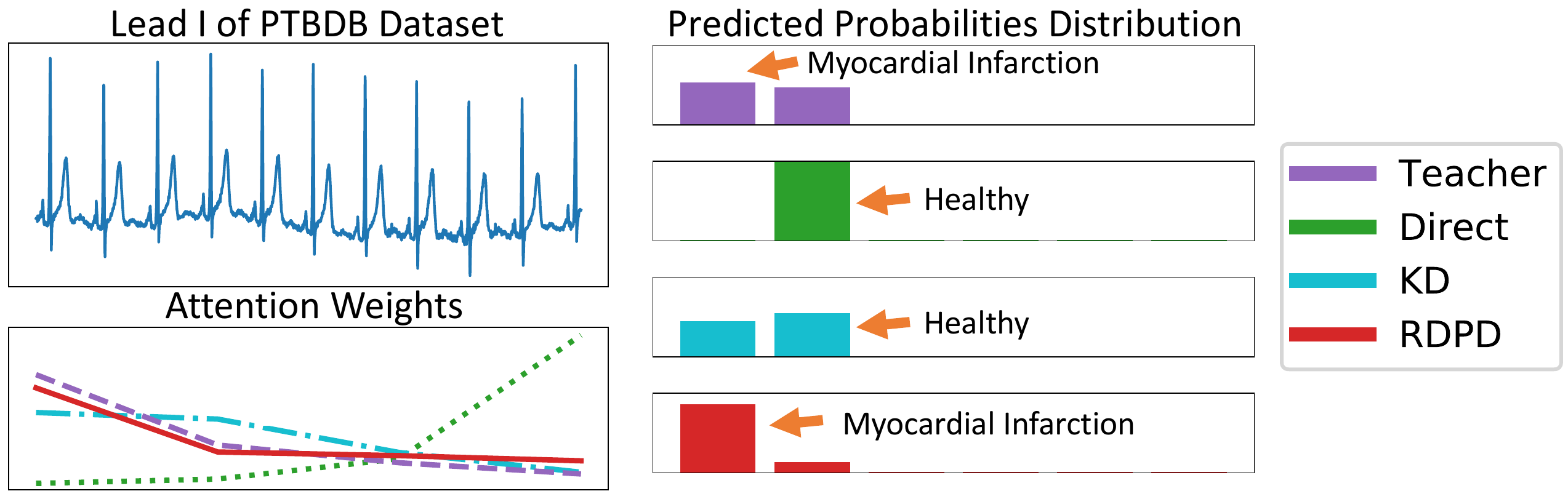}
\caption{An example of PTBDB data and model predictions. (Left Upper) Original Lead I ECG data. (Left Bottom) Attention weights of different methods. (Right) Predicted probabilities distribution of different methods. }
\label{fig:case_ptbdb}
\end{figure}

An example of PTBDB data and model predictions are shown in Figure \ref{fig:case_ptbdb}. \mname and Teacher give correct predictions while Direct and KD are wrong. The reason of \mname better than KD and Direct comes from two aspects. One the one hand, \mname and KD imitate Teacher's soft label so that they correct wrong predictions like Direct in some extent. On the other hand, \mname also imitate Teacher's attention weights (shown in Left Bottom, purple is Teacher and red is \mname), so that \mname gives more confident predictions of Myocardial Infarction than KD, and thus further correct final predictions given by KD. Besides, we also found that \mname gives more confident predictions of Myocardial Infarction than teacher model. The reason is that \mname also considers combined label, and it leverages the soft label from Teacher model and hard label thus would be more confident when soft label and hard label are consistently right.

\begin{figure}[ht]
\centering
\includegraphics[width=\linewidth]{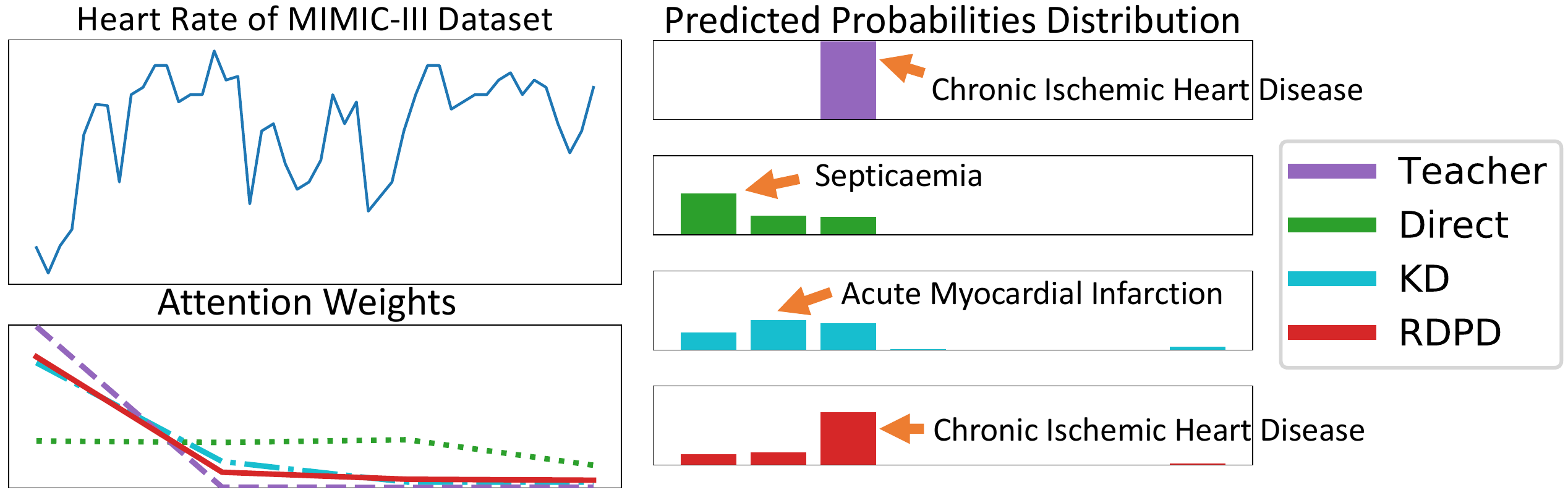}
\caption{An example of MIMIC-III data and model predictions. (Left Upper) Original heart rate data. (Left Bottom) Attention weights of different methods. (Right) Predicted probabilities distribution of different methods. }
\label{fig:case_mimic}
\end{figure}

An example of MIMIC-III data and model predictions are shown in Figure \ref{fig:case_mimic}. \mname and Teacher give correct predictions while Direct and KD are wrong. When looking at the attention weights (shown in left bottom), the Direct shows average weights on all part, while Teacher, KD and \mname emphasize at the beginning. Moreover, in the first and the second part of attention weights, \mname are in the middle of KD and Teacher, which reveal that \mname successfully learns Teacher's attentions that changes from KD to Teacher. In the right part, we can see that the probabilities of correct prediction increases from Direct to KD, and further increases in \mname. 

\section{Effect of Temperature}

\begin{figure}[ht]
\centering
\includegraphics[width=\linewidth]{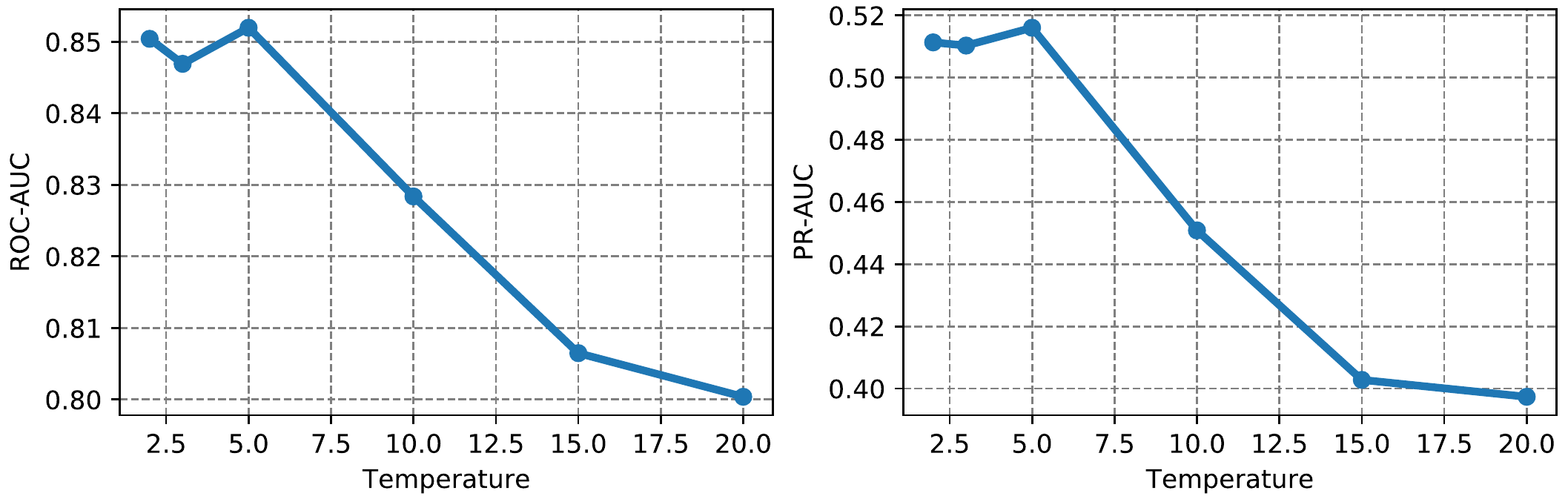}
\caption{Performance comparison of temperature $T$ in \mname on PAMAP2 dataset. (Left) ROC-AUC, (Right) PR-AUC. }
\label{fig:temperature}
\end{figure}

Temperature $T$ is one critical hyperparameter that controls the degree of smoothness for soft labels. The larger of smoothing parameter $T$, the smoother of probability distribution over classes. 
To determine a proper value for $T$, our decision was based on the ROC-AUC and PR-AUC in validation set.
In Figure \ref{fig:temperature} we plot the effects of hyperparameter $T$. It is easy to see that both ROC-AUC and PR-AUC slightly increase along the of $T$  from $1$ to $5$, then start to drop as $T$ becomes larger. The reason is that larger $T$ leads to a softer probabilities distribution, thus the model would be less discriminative and consequently yield lower ROC-AUC and PR-AUC. In our experiments, we set $T=5$ on PAMAP2 dataset as a proper choice.

}

\end{document}